\documentclass[12pt]{article}

\usepackage{times}
\usepackage{hyperref}
\usepackage{algorithm2e}
\usepackage{graphicx}
\usepackage{amssymb}
\usepackage{amsfonts}
\usepackage{mathtools}
\usepackage{verbatim}
\usepackage{listings}
\usepackage{color}
\usepackage{makecell}
 
\definecolor{codegreen}{rgb}{0,0.6,0}
\definecolor{codegray}{rgb}{0.5,0.5,0.5}
\definecolor{codepurple}{rgb}{0.58,0,0.82}
\definecolor{backcolour}{rgb}{0.95,0.95,0.92}
 
\lstdefinestyle{mystyle}{
    backgroundcolor=\color{backcolour},   
    commentstyle=\color{codegreen},
    keywordstyle=\color{magenta},
    numberstyle=\tiny\color{codegray},
    stringstyle=\color{codepurple},
    basicstyle=\footnotesize,
    breakatwhitespace=false,         
    breaklines=true,                 
    captionpos=b,                    
    keepspaces=true,                 
    numbers=left,                    
    numbersep=5pt,                  
    showspaces=false,                
    showstringspaces=false,
    showtabs=false,                  
    tabsize=2
}
 
\newcounter{lastnote}

\title{Realizing Continual Learning through Modeling a Learning System as a Fiber Bundle} 

\author
{Zhenfeng Cao$^{1\ast}$\\
\\
\normalsize{$^{1}$Department of Computing, The Hong Kong Polytechnic University, KL, Hong Kong}\\
\normalsize{$^\ast$To whom correspondence should be addressed; E-mail:  zhenfengcao@yeah.net}
}

\date{}

\begin{document} 


\baselineskip24pt


\maketitle 

\begin{abstract}
A human brain is capable of continual learning by nature; however the current mainstream deep neural networks suffer from a phenomenon named catastrophic forgetting (i.e., learning a new set of patterns suddenly and completely would result in fully forgetting what has already been learned). In this paper we propose a generic learning model, which regards a learning system as a fiber bundle. By comparing the learning performance of our model with conventional ones whose neural networks are multilayer perceptrons through a variety of machine-learning experiments, we found our proposed model not only enjoys a distinguished capability of continual learning but also bears a high information capacity. In addition, we found in some learning scenarios the learning performance can be further enhanced by making the learning time-aware to mimic the episodic memory in human brain. Last but not least, we found that the properties of forgetting in our model correspond well to those of human memory. This work may shed light on how a human brain learns.
\end{abstract}
\section{Introduction}
Recent progress in neuroscience indicates that our brain is a continual learner by nature \cite{hayashi2015labelling,cichon2015branch}. Continual learning is crucial, especially in scenarios of learning from streaming data or that the searching space is unusually large (e.g., in many reinforcement-learning scenarios), in that it can reduce the loads of relearning what has already been learned in the past and hence effectively increase the learning speed without sacrificing the stability of learning. In addition, this capability may also benefit meta-learning\cite{santoro2016meta} and transfer learning\cite{pan2010survey}, in which a previously trained model is required to solve a new problem through a quick adaptation to a new learning environment. However, current mainstream neural networks (NNs) have very limited capability of continual learning due to catastrophic forgetting\cite{mccloskey1989catastrophic,ratcliff1990connectionist}. Catastrophic forgetting is also known as the stability-plasticity dilemma\cite{mermillod2013stability}, a well-known constraint for artificial and biological neural systems. Existing solutions to the problem of catastrophic forgetting mainly involve introducing strategically experience replay\cite{mnih2015human,shin2017continual,parisi2018continual}, and evaluating dynamically the importance of the trainable parameters by calculating their Fisher matrices when updating them\cite{kirkpatrick2017overcoming,snell2017prototypical,amari1998natural,finn2017model}. However, experience replay would slow down the learning process because extra time are spent on relearning what has already been learned in the past; and calculating the Fisher matrices of the training parameters are often not computationally efficient for large networks and is also not easy to implement for arbitrary neural-network architectures. Most importantly, existing solutions are quite different from how a human brain learns: our brain learns through analogy, interpreting and understanding new situations in terms of prior experience (i.e., What does this look like? Have I experienced similar situations?)\cite{bar2007proactive}. In addition, our experiences are represented in structures that cluster together related information. For instance, objects that tend to appear together are linked on some level; and the representations of them include properties that are inherent to and typical of that same experience\cite{bar2007proactive}. Such structures are termed ``context frames'' in neuroscience\cite{bar2007proactive,bar1996spatial,bar2004visual}. Biologically, it is found that a learning from a new task depends on the formation of a ``task-specific'' dense synaptic ensemble\cite{hayashi2015labelling}.

Borrowing inspiration from differential geometry and the latest development in cognitive science and neuroscience, in this work we model a learning system as a fiber bundle, which is able to capture naturally both strong and weak correlations and encode objects according to certain similarity measure emerged from evolution (i.e., learning) and hence enables automatically continual learning. In addition, we introduce a context frame by making the model time-aware through an adaptive bioclock. By doing so, the effect of catastrophic forgetting is further reduced. 

The paper is organized as follows. We first introduce the theories, and then do a variety of experiments to study the properties of our models. In addition, the correspondences between the properties of our models and those of a human brain found in cognitive science and neuroscience are discussed in detail. Extended discussions as well as a conclusion are provided at the end. 

\section{Theory}
\begin{figure}[!ht]
  \centering
  \includegraphics[width=0.8\textwidth]{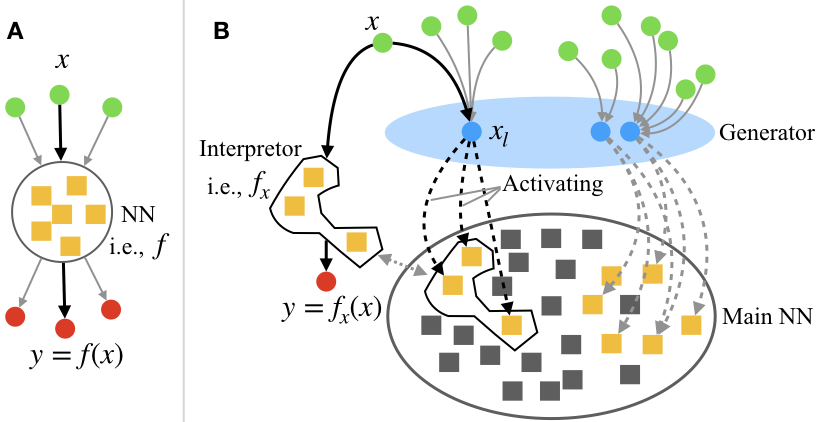}
  \caption{a. A sketch showing the architecture of a MLP, where a single interpretor $f$ is used for interpreting all inputs. b. A sketch showing the architecture of our proposed learning model; a dashed arrow from node A to square B means ``A activates B'', the gray dotted double arrow means ``equivalent'', and the blue nodes (i.e., $x_{l}$) are objects in the latent space; the generator generates an interpretor $f_{x}$ for interpreting $x$ by selectively activating some of the neurons in the main neural network. In both figures, green nodes represent inputs, red nodes represent outputs, solid arrows show how information flows, and squares represent neurons (gray ones are inactivated, and yellow ones are activated).}
\end{figure}
Here we introduce our model by following the learning process of a human brain. A main feature of how a brain learns is that when it receives a piece of new information it will automatically think about if it has learned similar things already in the past\cite{bar2007proactive}, and then selectively activates neurons to form an interpretor most suitable for interpreting this new information\cite{snodderly2001selective,holtmaat2016functional} (Fig.1b). The current mainstream neural networks' learning processes, however, process all information with a single (Fig. 1a) or a few manually selected interpretors, and are often not able to ``automatically'' select the best interpretor basing on analogy. In addition, conventionally, we model the neural network and observations separately, here we are modeling them together as a self-contained learning system since the learning could be viewed as a mapping of objects in the environment into our mind. We found that with all these considerations, the learning system corresponds well to an adaptive smooth fiber bundle. We require it to be smooth so that the model can be trained using backpropagation. 

Formally, a fiber bundle is defined as follows\cite{rowland2016fiber}.  A fiber bundle with fiber $F$ is a map $f:E\to B$, where $E$ is called the total space of the fiber bundle and $B$ the base space of the fiber bundle. The main condition for the map to be a fiber bundle is that every point $b \in B$ has a neighborhood $U$ such that $f^{-1}(U)$ is homeomorphic to $U\times F$ in a special way. Namely, if $h:f^{-1}(U) \to U \times F$ is the homeomorphism, then $\textrm{proj}_{U} \circ h=f_{|f^{-1}(U)|}$, where the map $\textrm{proj}_U$ means projection onto the $U$ component. The homeomorphisms $h$ which ``commute with projection'' are called local trivializations for the fiber bundle $f$. In other words, $E$ looks like the product $B \times F$ (at least locally), except that the fibers $f^{-1}(x)$ for $x \in B$ may be a bit ``twisted'' \cite{steenrod1999topology,rowland2016fiber}. When the range of a function (i.e., an interpretor) only makes sense locally, it is necessary to use bundles\cite{rowland2016fiber}. With this knowledge, we suggest a fiber bundle is very suitable for modeling a continual learning system in that an interpretor may only be valid for interpreting a subset of data belonging to the same task and a bundle of interpretors are, however, potentially capable of interpreting data from multiple learning tasks.

Figure 1b shows a sketch for our proposed learning model. The model 1) reads an input  sample $x$ from the observation space $X$, 2) encodes it into an object $x_{l}$ in a latent space $L$, in which similar objects are encoded similarly, 3) selectively activates (mathematically, this may correspond to ``assigns large weights to'') neurons to form dynamically an interpretor $f_{x}$ basing on $x_{l}$, 4) and then interprets $x$ using $f_{x}$ to get the output $y$. The correspondence of our model to a fiber bundle is as follows: an interpretor corresponds to a point (or object) in the base space $B$, and the subset of samples that are most suitable to be processed by this interpretor form the fibers rooted at this point, and $B\times F$ forms the total space $E$ (which is a trivial fibration). The maps of  $X$ onto $L$ and then onto $B$ are all surjective, so that each point in $B$ could correspond to many input samples. The fiber bundle that achieves the goals the best would automatically emerge from learning.

Next we introduce a practical basic architecture. Since a fully connected linear layer (i.e., a single-layer perceptron) is a building block of a large number of existing neural networks, we modify it to obtain a minimum version of our model, so that the readers can construct more delicate ones with it or following the concept of it. Practically, it is convenient to represent all objects in vectors and matrices; therefore, here we assume all spaces are vector spaces. First, we define a linear transformation: $L(X; W, B)=XW+B$, where $W$ and $B$ are a weight matrix and a bias vector, respectively. Then a \emph{simple bundle layer} is constructed as follows.
\begin{equation}\label{eqn:simple-bundle}
\begin{split}
G &= \sigma [L(X; W_{g}, B_{g})],\\
X_{l} &= L(G*X; W_{l}, B_{l}),\\
W &= \tanh[L(X_{l}; W_{w}, B_{w})],\\
B &= L(X_{l}; W_{b}, B_{b}),\\
Y &= L[(1-G)*X; R(W), B],
\end{split}
\end{equation}
where $X$ is the input, $Y$ is the output, $\sigma$ is the sigmoid function, ``$*$'' is the entry-wise product, $G$ is a gate controlling the flows of $X$, entries of any ``$W$'' or ``$B$'' with a subscription are trainable parameters, and $R$ is a reshaping function that reshapes $W$ to a proper shape to enable the matrix product of $(1-G)*X$ and $W$. $G$, $B$ and $\tanh$ are for stabilizing the training, and are optional (especially when the input and output data are well bounded). The key concept to construct a bundle layer is that we make the parameters of an interpretor as the output of another neural network whose input involves $X$.

To make the learning time-aware, we introduce a \emph{bioclock}, which is given by
\begin{equation}\label{eqn:bioclock}
\begin{split}
T &= T_{min}+\sigma[\psi(X)]*(T_{max}-T_{min}),\\
Y_{t} &= (A\sin(2\pi t/T), A\cos(2\pi t/T)), 
\end{split}
\end{equation}
where $t$ is the current time (e.g. the step number in an episode in some reinforcement-learning tasks), $X$ is the input data, $Y_{t}$ is the output of the bioclock (i.e., encoded time), $T$ is the period, $T_{min}$ and $T_{max}$ are hyper-parameters, and are the minimum and maximum possible values of $T$, respectively, $A$ is the amplitude, which is a trainable parameter, $\sigma$ is the sigmoid function, and $\psi$ is a trainable neural network. There are two advantages to use periodic functions to construct a bioclock: 1) the output of it is well bounded, and 2) the non-periodic cases could be regarded as the case when $T$ is sufficiently large. Next, we do a series of experiments to study the properties of learning systems with a fiber bundle topology and the effect of introducing a bioclock. 
\section{Experiment}
\subsection{Learning speed and information capacity}
\begin{figure}[!ht]
  \centering
  \includegraphics[width=0.8\textwidth]{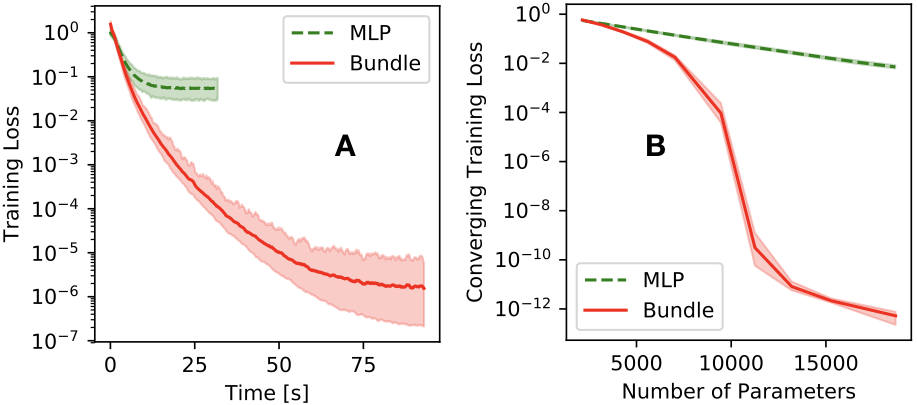}
  \caption{a. The evolution of the training loss of a Bundle vs. that of a MLP; x-axis shows the elapsed walk-clock time; we run both models for 2000 epochs to ensure that the trainings are roughly converged. b. This figure studies the per-parameter information capacity of both models by comparing the converging training losses of both models when the numbers of trainable parameters are the same. In both Fig. a and b, the error bands in light colors represent the minimum and maximum values of 20 repeated runs.}
\end{figure}
For convenience of discussion, hereafter we name a learning system with a fiber bundle topology by ``Bundle'', and one with a conventional neural network formed by multilayer perceptrons by ``MLP''. In all experiments we ensure that the number of trainable parameters in a Bundle is $\lessapprox$ that in a MLP, and all other settings are controlled to be the same. In Fig. 2, we stack two \emph{simple bundle layers} (i.e., Eq. \ref{eqn:simple-bundle}) with a $\tanh$ activation on the output of the first layer to form the NN of a Bundle, and stack two fully connected perceptrons (i.e., linear layers) with a $\tanh$ activation on the output of the first layer to form the NN of a MLP. The trainings in Fig. 2 are of offline fashion (i.e., allowing repeatedly training on the same but shuffled dataset). In Fig. 2a, we generate a fixed number of pairs of input and output vectors whose entries are values randomly drawn from a standard normal distribution; then we train a Bundle and a MLP on these vector pairs for a sufficiently large number of epochs and compare the evolutions of their training loss. From this figure we can see that although the total time taken by the Bundle to complete the same number of epochs is around four times of that of the MLP (in real implementations it is usually sufficient to replace only some layers of a conventional deep NN by simple bundle layers; hence ``four times'' is an upper bound), to achieve the same level of loss the Bundle takes much less time. To study the information capacity, we prepare the dataset as that for Fig. 2a, and compare the relationships between the converging training losses (i.e., the training losses measured when the trainings are roughly converged) and the numbers of trainable parameters of both models. The number of trainable parameters for each model is adjusted by varying the size of the output of the first layer. We can see from Fig. 2b that the converging training loss of the Bundle decreases much faster than that of the MLP with the increase of the number of trainable parameters, indicating that a Bundle has a larger per-parameter information capacity than a MLP. This is understandable because the Bundle automatically ``groups'' similar samples and interpret them with a most suitable interpretor, while the MLP interprets all samples with a single interpretor. A Bundle's superior information capacity becomes significant especially when the dataset is highly irregular (e.g., the representations of the samples are highly clustered). 
\subsection{Continual learning and catastrophic forgetting}
\begin{figure}[!ht]
  \centering
  \includegraphics[width=0.8\textwidth]{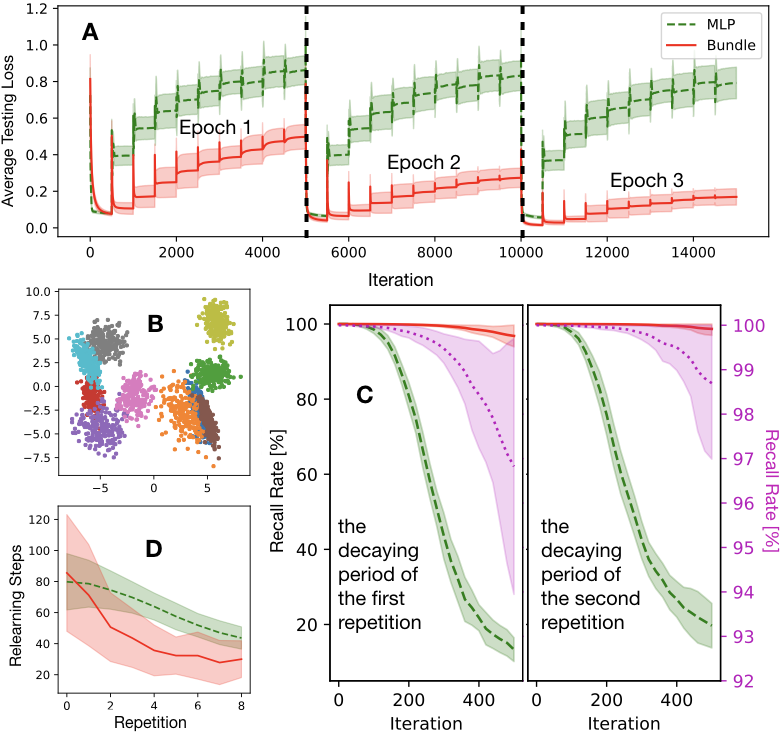}
  \caption{a. The evolution of the average testing loss of a Bundle vs. that of a MLP when training sequentially on 10 different tasks in every training epoch; the average testing loss measured when completing the current task in every epoch is defined as the average of the testing losses of all the already learned tasks (including the current one) in that epoch; we repeat the training by three epochs to show how the two models make progress if we allow the model to relearn what have been learned for a few times. b. A visualization of the PCA of $W$ (defined in Eq. \ref{eqn:simple-bundle}) of the first layer of the Bundle calculated by inputting a subset of testing data; each dot corresponds to an interpretor; interpretors for different tasks are distinguished by colors. c. The evolution of the recall rates of both models in the decaying periods of the first (i.e., right after the initial learning period) and the second repetition. d. The number of relearning steps needed to recover the original recall rate (i.e., 100\%) vs. the number of repetitions. The legends in Fig. a is applicable to Fig. c and d. In Fig. d, we provide an enlarged view of the evolution of the recall rate of the Bundle on the right-hand-side y-axis. The error bands in light colors in Fig. a, c, and d represent the standard deviations.}
\end{figure}
Next we compare the learning performance of a Bundle and a MLP in an online-learning setting (i.e., allowing limited or no access to previously trained datasets when training on the current dataset), in which the capability of continual learning becomes crucial. Specifically, we let both a Bundle and a MLP learn from 10 different tasks ``sequentially'' in every epoch. The NN of the MLP for this experiment is constructed by stacking two linear layers with a $tanh$ activation on the output of the first layer; and that of the Bundle models is constructed in the same way except that we replace the first linear layer of the MLP model by a simple bundle layer. For each task we generate a training dataset and a testing dataset. A sample $(x, y)$ in these datasets is generated as follows: we first generate randomly a transformation matrix $M$ and a feature vector $v$, both of which are unique and fixed for every task; and then we generate an input vector $x$ by concatenating a random vector $x'$ and $v$; next, we generate an ouput vector $y$ for this input by multiplying $x$ by matrix $M$. A task is then defined as finding an approximation to this transformation. The entries of all these vectors and matrices are values drawn from the standard normal distribution. $v$ serves as a feature allowing the model to distinguish the tasks, analogous to a ``context frame'' in neuroscience\cite{bar2007proactive,bar1996spatial,bar2004visual}. For each task, we train the models for a sufficiently long period before moving to the next one to ensure the current task has been sufficiently learned. The (average) testing loss vs. the number of iterations are shown in Fig. 3a. It shows that the MLP forgets almost completely what has been learned in previous tasks after training on a new one, while the Bundle suffers much less from this catastrophic forgetting. Furthermore, if we allow the models to relearn a few times (corresponding to the 2rd and 3rd epochs in Fig. 3a) what has been learned, while the MLP still makes little progress, a significant improvement in performance has been observed in the Bundle, which indicates further the Bundle has a stronger continual-learning capability than the MLP. 

To understand why the Bundle has a stronger continual-learning capability than the MLP, we apply principal component analysis (PCA) to the weight matrix ($R(W)$ in Eq. \ref{eqn:simple-bundle}) of the first layer of the trained Bundle. Specifically, we generate the weight matrices by inputting a number of testing sample, flatten them to be vectors, and then apply PCA on them to reduce the numebr of dimensions to two for visualization (Fig. 3b). It is found that for each inputting sample the Bundle allocates a unique interpretor (i.e., a weight matrix) to interpret the sample. In addition, interpretors for the same task (shown is in the same color) are encoded automatically close to each other to form a cluster, and clusters corresponding to similar tasks are also close to each other. This is analogous to the phenomenon of learning based on analogy in cognitive science\cite{bar2007proactive}. In addition, this clustering and object-specific allocation of interpretors also provides a possible explanation for the object-selective activations of brain regions (e.g., different object categories are found to activate different regions across the ventral and dorsal streams\cite{snodderly2001selective}) as well as the evolutionary formation of brain's functional organization\cite{johnson2001functional,zeki1991direct,luria1970functional,gazzaniga1989organization,diez2015novel}. 

Next, we study the stability of memory by applying perturbations to the learned models. Specifically, we train a Bundle model and a MLP to do a classification task. We train both models fully so that the recall rates (i.e., the training accuracy) of them reach 100\%, and then perturb them by training them with random inputs and outputs for a fixed number of iterations. It is found that the recall rates of both models decay roughly exponentially after a short accelerating period (Fig. 3c), resembling the forgetting curve of human memory\cite{frankstarmer,ebbinghaus}. Besides, the decaying speed of the Bundle is found to be much lower than that of the MLP (Fig.3c). In addition, if we apply ``learning with spaced repetition''\cite{ebbinghaus,murre2015replication} (i.e., repeating ``relearning what has been learned again after a fixed period of decay'' several times), the models (especially the Bundle) become less prone to forgetting with the increase of the number of repetitions. This is further shown as that the final recall rates after the decaying period of the second repetition are higher than those of the first repetition (Fig. 3c), and that the relearning steps required to recover the original recall rate (i.e., 100\%) decreases with the number of repetitions (Fig. 3d). All these findings are consistent with the findings in the studies of human memory\cite{ebbinghaus}.
\subsection{Reinforcement learning}
\begin{figure}[!ht]
  \centering
  \includegraphics[width=0.8\textwidth]{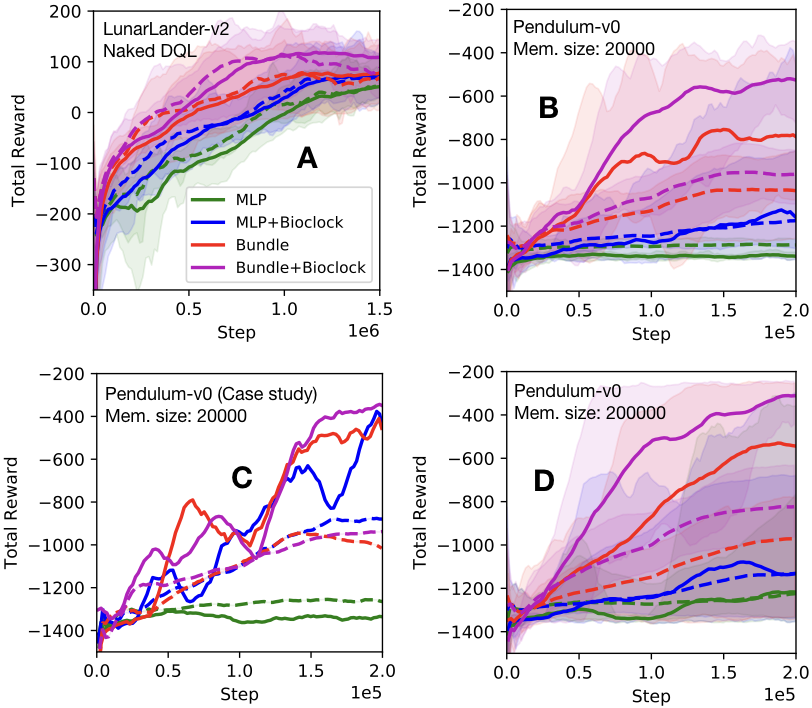}
  \caption{a. The evolutions of the testing (solid lines) and training (dashed lines of the same colors) total rewards achieved by our models (a Bundle with or without a bioclock) and the baseline ones (a MLP with or without a bioclock) when training on the LunarLander-v2 task using a naked DQL. b. Similar to Fig. a, but training on the Pendulum-v0 task with DDPG equipped with a small (20000 samples) memory buffer for experience replay but no target NN. c. A case study of experiments in Fig. b. d. The same with b and c, except that we use a large (200000 samples) memory buffer for experience replay. The legends in Fig. a is applicable to all figures. The error bands in light colors show the minimum and maximum values of the total rewards of 10 repeated runs.}
\end{figure}
Finally, we apply the simple bundle layer (Eq. \ref{eqn:simple-bundle}) as well as the bioclock (Eq. \ref{eqn:bioclock}) to two classic and nontrivial reinforcement-learning (RL) tasks in OpenAI Gym\cite{brockman2016openai}, named LunarLander-v2 and Pendulum-v0, and study how our proposed models help improve the learning performance. LunarLander-v2 has a continuous state space and a discrete action space, and Pendulum-v0's state space and action space are both continuous. Specifically, we use a RL algorithm named Deep Q-Learning (DQL)\cite{mnih2013playing,mnih2015human} that combines a learning agent formed by a deep NN with a well-known RL algorithm named Q-learning\cite{watkins1992q,sutton1998introduction} for the task of  LunarLander-v2 and a RL algorithm named DDPG\cite{lillicrap2015continuous} (which is conceptually similar to the previous one but in an actor-critic style and is for tasks whose action space is continuous) for the task of Pendulum-v0. Usually, it is necessary to introduce a slowly-updating ``target'' neural network\cite{lillicrap2015continuous} as well as a large memory buffer for experience replay\cite{mnih2013playing,mnih2015human} to ensure the stability of training. To make the experiments challenging, however, here we use a ``naked'' (i.e., no target network and no experience replay) version of DQL for the task of LunarLander-v2, and use only a memory buffer for experience replay (both a small one and a large one are studied for comparison) but no target network for the task of Pendulum-v0. In such a setting, the capability of continuous learning becomes important. Here are the details of our experiments and results.

In the task of LunarLander-v2, the NN for the MLP model is constructed by stacking two linear layers with the $tanh$ activation function followed by a layer normalization (for stabilizing the hidden state dynamics)\cite{ba2016layer} acting on the output of the first layer, and that for the Bundle model is constructed in the same way except that we replace the first layer by a simple bundle layer. The neural networks for the actor and the critic of both models for the task of Pendulum-v2 is constructed in the same way except that we add an extra $Tanh$ activation on the output of the actor to bound the action values. With respect to the task of LunarLander-v2, since experience replay is not allowed, we train the sample batches sequentially (i.e., no access to previously trained sample batches) by multiple forwarding and backpropagation steps to ensure that the current batch is well learned before moving to the next one. With respect to the task of Pendulum, we equip the algorithm with a small (whose volume is 20000 samples) or a large (whose volume is 200000 samples) memory buffer for experience replay to study the influence of the size of the memory buffer, and forward and backpropagate by one step for every exploration step. For any model with a bioclock, the bioclock is constructed by setting $\psi$ in Eq. \ref{eqn:bioclock} to a simple bundle layer. We concatenate the output of the bioclock (i.e.,  $Y_{t}$ in Eq. \ref{eqn:bioclock}) with the conventional observation state $S_{t}$ to form a full input state vector. It is found that in all cases the Bundles learn faster and more stable than the MLPs, and introducing a bioclcok can further enhance the learning performance for both the MLPs and the Bundles (Fig. 4). 

To understand why a Bundle performs better than a MLP, we present a case study for the Pendulum-v0 task (Fig. 4c), it is found that the evolutions of the total reward for all models show roughly a pseudo-periodic increasing-decaying pattern, the duration of whose increasing phase is roughly equal to the size of the memory buffer. This indicates when the number of learning steps goes beyond the volume of the memory buffer learning new samples would result in quick forgetting of previously learned ones in the MLPs; by contrast, the Bundles perform more stably by naturally retaining information remains useful and learning based on prior learning experiences. This becomes possible since for every input sample the Bundle selectively allocates a most suitable interpretor (which could be regarded as a weighted combination of trainable parameters) as the initial one for further improvement, so that the parameters relevant for constructing this interpretor are effectively trained (errors can propagate effectively back to highly weighted parameters), while those less relevant are less trained. This is analogous to what happens in human learning. For instance, mastering the programming language Matlab may benefit the learning of Python since they are quite similar to each other; however it is less likely that mastering Matlab would help the learning of painting since they are quite different from each other. However, although in our studied cases a Bundle performs better in learning than a MLP, a good memory doesn't always mean a high learning capability for a new task: 1) for very simple learning tasks, a MLP may learn slightly quicker since the training is more ``focused'' (i.e., training a single interpretor on all samples), and 2) when information in a new task contradict that in the old ones, prior experience may impede the learning of the new one, the effect of which may be more significant in a Bundle than in a MLP since a Bundle learns quickly but forgets slowly, while a MLP learns and forgets both quickly.  Hence, we suggest that a Bundle is not for replacing a MLP---they are just like long-term and short-term memory modules in human brain, each of which has its own pros and cons. So, when designing a learning system, we can combine them properly to exploit their distinct capabilities.  

With respect to the influence of a bioclock (i.e., time-awareness), the reason why it helps learning in some scenarios is that it provides a context frame, which is crucial for the formation of long-term memory (more precisely, episodic memory)\cite{tulving1972episodic,tulving1973encoding}. Imagine that if there is no time label for our memory, our personal experiences would be superposed so that it would be hard to distinguish them. However, introducing a bioclock doesn't always improve the efficiency and stability of learning since it introduces extra information and, to cetain extent, defocuses the training. Hence, we need to evaluate case-by-case if it is necessary or better to make the learning time-aware. 

Last but not least, it is found that a large memory buffer for experience replay doesn't guarantee improvement in learning performance especially for a MLP if we compare the performances of MLPs in Fig. 4b and 4d. Sometimes it is even better to be able to forget. The reasons are at least two folded: 1) the information capacity of a NN is always finite and hence when the memory buffer is large and new samples keep coming, the NN may run out of information capacity soon; and 2) not all experiences are equally worth memorizing given that the time for learning is limited and hence selective forgetting outdated experiences and retaining only useful ones would make the learning more focused and hence lead to faster progress.

\section{Conclusion and outlook}

In conclusion, in this work we highlighted the role of topology of a learning system in influencing the continual-learning capability. Specifically, we showed that a fiber bundle can effectively model both strong and weak correlations and enables naturally continual learning. In addition, we found the learning and memory properties of our model correspond well to those of a human brain observed in cognitive science and neuroscience. However, there are still many open questions. For instance, our understanding of the biological realization of the mechanisms indicated by our mathematical model remains unclear. Besides, a fiber bundle is a very flexible and highly abstracted geometric object, our modeling is still very naive: its learning capability could be further improved by incorporating more latest advances in cognitive science and neuroscience. Moreover, our model exhibits an interesting property of ``easy to memorize but hard to forget'', which is very similar to the property of the formation of long-term memory in human brain. Given that many psychological diseases are associated with memory about suffering experiences, it would therefore be interesting to explore based on our model how we can selectively and quickly forget unwanted experiences while retaining wanted ones. In addition, considering the enhanced continual-learning capability and high information capacity of a Bundle, a possible future direction might be on building a learning model with multiple highly specialized senses (e.g., vision, hearing, taste, smell, and touch) and allow information from different sources to interact with each other naturally, which might enhance the learning in that they provide context frames for each other. Finally, though a fiber bundle has been suggested in this work as a possible topology of the human learning system, it remains an open question if there is any better alternatives. This work may serve as a stepping stone for narrowing the gap between a realistic mathematical description of our brain's learning process and the various theoretical hypotheses and biochemical observations in cognitive science and neuroscience on how our brain learns .

\section*{Acknowledgments}
The author would like to thank for the financial support from the department of computing of the Hong Kong Polytechnic University as well as the helpful advices from Prof. Jiannong Cao.
\bibliography{bundle}

\bibliographystyle{ieeetr}

\newpage
\section*{Supplemental Materials}
Here we introduce the technical details of the experiments whose results are shown in the figures of the main text. We did all the experiments using PyTorch v0.4.1 (availabel at \url{https://pytorch.org/})\cite{paszke2017automatic}. For all experiments, linear layers (including those inside the simple bundle layers) are constructed using the $torch.nn.Linear$ module in PyTorch. For experiments in Fig. 2-3, The weight matrices and biases of the linear layers are initialized by the default initialization of the $torch.nn.Linear$ module. For experiments in Fig. 4, the weight matrices of the linear layers of the main NNs of both the MLPs and the Bundles as well as the bioclocks are initialized using the $torch.nn.init.orthogonal\_$ with its parameter $gain=5/3$; the biases of them are initialized to be zero; and other linear layers are initialized by the default initialization of Pytorch. The ``layer normalization'' layers are implemented using the $torch.nn.LayerNorm$ module in PyTorch, and are initialized by the default initialization of the module. For any other trainable parameters unmentioned, we initialize them using values sampled from the standard normal distribution. In addition, all optimizations are done using the Adam optimizer\cite{kingma2014adam}, and all experiments are done on a server with 24 CPUs and a NVIDIA Quadro P4000 GPU.

\subsection*{Technical details of Fig. 2}
The structures of the NNs, the objective of training, as well as the generation of the datasets are introduced in detail in the main text. Here are the detailed settings: for both Fig. 2a and 2b, the number of samples is 200, the size of the input and the ouput are both 16, the size of a batch is 20, the total number of epochs is 2000, and the total number of repeated runs is 20; the mean squared error of the predicted output and the observed output is used as the loss function, the learning rate of epoch $t$ is $0.001*0.995^{t}$; for Fig. 2a, the size of the hidden state (i.e., the output of the first layer) of the MLP is 323 (so that the total number of trainable parameters is 10675), the size of the hidden state of the Bundle is 16 (so that the total number of trainable parameters is 10336); for Fig. 2b, we vary the size of the hidden state of the Bundle from 4 to 24, and that of the MLP from  64 to 565; and when making Fig. 2a , we smoothen the curves using a simple moving average whose window size is 50. 

\subsection*{Technical details of Fig. 3}
The NN of the MLP for this experiment is constructed by stacking two linear layers with a $tanh$ activation on the output of the first layer; and that of the Bundle models is constructed in the same way except that we replace the first linear layer of the MLP models by a simple bundle layer. When doing experiments in Fig. 3c and 3d we add an extra \emph{LogSoftmax} (i.e., the logarithm of the softmax function) activation layer on the final output, and use the negative log-likelihood loss as the loss function. 

\subsubsection*{Technical details of Fig. 3a and 3b}
The procedure of data preparation for Fig. 3a is introduced in detail in the main text.  We train the model in the same way with that in Fig. 2. Other settings are as follows: the size of the input is 16 (i.e., an input sample is constructed by concatenating a random vector $x'$ whose size is 8 and a task-specific feature vector $v$ whose size is also 8), the size of the output is 8, the size of the hidden state of the MLP is 220 (so that the total number of trainable parameters is 5580), the size of the hidden state of the Bundle is 16 (so that the total number of trainable parameters is 5304), the size of a batch is 50, the number of iterations for a task in each run is 500, and the number of repeated runs is 20. When making Fig. 3b, we use the module \emph{sklearn.decomposition.PCA} of a Python package named $sklearn$\cite{scikit-learn} to apply PCA on the flattened $W$ ( defined in Eq. \ref{eqn:simple-bundle}) of the first layer of a Bundle model trained on the tasks in Fig. 3a. to reduce the dimension to two. 

\subsubsection*{Technical details of Fig. 3c and 3d}
The datasets in Fig. 3c and 3d are generated as follows: an input vector $x$ is generated in the same with that in Fig. 3a, however, its corresponding output becomes an integer labeling which task this input vector belongs to. We generate many input and output pairs for each task, and then mix them for different tasks to form a full dataset. Then the objective becomes classifying these input vectors by tasks. We train the models by ``spaced repetition'', which means in each repetition we 1) train the models with the data samples until they converge (i.e., the recall rates become 100\%), and then 2) train them with noises (i.e., random input and output) for a fixed number of iterations.We call the first training period in a repetition the ``learning period'', and the second training period in the repetition the ``decaying period''.  In Fig. 3c we study the recall rate vs. time (i.e., training steps) during the decaying periods, and in Fig. 3d, we look at the number of relearning (i.e., retraining) steps needed to recover the original recall rate (i.e., 100\%) vs. the number of repetitions. 

For both Fig. 3c and 3d, the detailed settings are as follows: the number of samples for a class (i.e., a task) is 500, the size of the input is 16, the size of the output as well as the total number of classes (i.e., tasks) are both 20, the size of the hidden state of the MLP is 155 (so that the total number of trainable parameters is 5755), the size of the hidden state of the Bundle is 16 (so that the total number of trainable parameters is 5508), the maximum number of iterations for learning period is a sufficiently large number so that the recall rate can always recover to 100\% (here we set it to 501), the number of iterations for the decaying period in a repetition is 501, the size of a batch is 100, and the number of repeated runs is 20. When making the figures, we slightly smoothen the curves by a simple moving average whose window size is 2 for Fig. 3c and 4 for Fig. 3d.

\subsection*{Technical details of Fig. 4}
We firstly briefly introduce the algorithms of Deep Q-Learning\cite{mnih2013playing} and DDPG\cite{lillicrap2015continuous}, and then introduce in detail how we modified these algorithms for our studies and how we did the experiments.

To start with, let's consider a typical reinforcement learning setup: an agent interact with an environment $E$. At each time step $t$, the agent takes an action $A_{t}$ with a probability calculated from some policy $\pi$ that maps the current state $S_{t}$ to a probability distribution over all possible actions, and then transits to another state $S_{t+1}$ and receives a reward $R_{t}$ that is a function of $S_{t}$ and $A_{t}$. Then the discounted total reward can be expressed as $G_{t} = \sum_{i=t}^{t_{m}} \gamma^{i-t} R_{i}$, where $t_{m}$ is the maximum time step, $\gamma $  ($\gamma \in [0,1]$) is the discount rate ensures $G_{t}$ is always finite and weights more on early rewards. The objective of learning is therefore to learn a best policy $\pi^{*}(A_{t}|S_{t})$ that maximizes the expected total reward when starting from state $S_{t}$ and taking an action $A_{t}$: $\pi^{*}(A_{t}|S_{t}) =  \mathop{\arg \max}_{\pi} Q^{\pi}(S_{t}, A_{t})=\mathop{\arg \max}_{\pi}  \mathbb{E}_{\pi}[G_{t}| S_{t}, A_{t}]$, where $Q^{\pi}$ is called the action-value function that describes the expected return after taking an action $A_{t}$ in state $S_{t}$ and following policy $\pi$. Optimal policies also share the same optimal action-value function, denoted $Q^{*}$, and defined as $Q^{*}(S_{t}, A_{t})=\max_{\pi} Q^{\pi} (S_{t}, A_{t})$. Instead of directly optimizing $Q^{\pi}$, it is more convenient to use the recursive relationship of it, which is known as the Bellman equation: $Q^{\pi}(S_{t}, A_{t})=\mathbb{E}_{R_{t}, S_{t+1}\sim E}\{R_{t} + \gamma \mathbb{E}_{A_{t+1}\sim \pi} [Q^{\pi}(S_{t+1}, A_{t+1})]\}$. When the target policy is deterministic we can describe it using a function $\mu$ that maps a state to a best action, and avoid the inner expectation: $Q^{\mu}(S_{t}, A_{t})=\mathbb{E}_{R_{t}, S_{t+1} \sim E}\{R_{t} + \gamma Q^{\mu}[S_{t+1}, \mu(S_{t+1})]\}$, which depends only on the environment. In Q-learning, we use a greedy policy to get the best action: $\mu(S_{t}) = \mathop{\arg \max}_{a} Q(S_{t}, A_{t})$. If we further use a deep neural network whose trainable parameters are $W^{Q}$ to approximate the $Q$ function, it becomes the so-called Deep Q-learning. Practically, the loss function of it is given by $L(W^{Q}) = \mathbb{E}_{S_{t}\sim \rho^{\beta}, A_{t}\sim \beta, R_{t}\sim E}\{[Q(S_{t}, A_{t}|W^{Q})-y_{t}]\}^2$, where $\beta$ is a policy for exploration (e.g., the current best policy plus some noise),  $\rho^{\beta}$ is the discounted state visitation distribution under policy $\beta$, and $y_{t}=R_{t} + \gamma Q(S_{t+1}, \mu(S_{t+1})|W^{Q})$. 

Conventional Deep Q-learning\cite{mnih2013playing} often involves a memory buffer for replay but has no time-awareness. Here we propose a time-aware Deep Q-learning algorithm without experience replay. The algorithm is shown in Algorithm \ref{algo1}, and is almost the same with Algorithm 1 in Ref. \cite{mnih2013playing}. There are only two main differences: 1) instead of sampling from a large memory buffer in every training step, we set the size of the memory buffer to be the size of a batch, train the batch when the buffer is full for multiple steps, and clear the buffer after this batch is trained; 2) for the cases with a bioclock, we store also the time label $t$ for each step in an episode and encode it using a bioclock as a state variable.

\begin{algorithm}
Initialize a memory buffer $M$ to capacity that equals to the size of a batch, $N_{b}$\\
Initialize an action-value function $Q(s, t, a; W^{Q})$ with random weights $W^{Q}$ (The bioclock is included in $Q$)\\
\For{episode=1, $N$}{
Initialize the environment to get the initial state $s_{1}$\\
\For{$t=1, t_{m}$}{
With a probability $\epsilon$ select a random action $s_{t}$, otherwise select $s_{t}=\max_{a} Q^{*}(s_{t}, t, a; W^{Q})$\\
Execute action $a_{t}$ in emulator and observe reward $r_{t}$ and the next state $s_{t+1}$\\
Store transition $(s_{t}, a_{t}, r_{t}, s_{t+1}, t)$ in $M$\\
\If{$M$ is full}{
Stack all the samples  in $M$ to form a batch $\{(s_{j}, a_{j}, r_{j}, s_{j+1}, t_{j})\}$ for training\\
\For{$\tau=1, \tau_{m}$}{
Set $y_{j}= r_{j} + \gamma \max_{a} Q(s_{j+1}, t_{j}+1, a;W^{Q})$ for non-terminal $s_{j+1}$, or $y_{j}=r_{j}$ for terminal $s_{j+1}$\\
Perform a gradient descent step to minimize the loss: 
$$L=\frac{1}{N_{b}}\sum_{j}^{N_{b}}[y_{j}-Q(s_{j},t_{j}, a_{j};W^{Q})]^2$$\\
}
}
}
}
\caption{Time-aware Deep Q-learning without experience replay. To disable the time-awareness, we just need to omit the bioclock and ignore the time labels. }\label{algo1}
\end{algorithm}

With respect to DDPG\cite{lillicrap2015continuous}, the basic concept of it is similar to that of Deep Q-learning. The main difference is that it is an actor-critic method: it uses an actor NN to read the current state and generate an action $a_{t}$, and then uses a critic NN to evaluate the ``quality'' of ``taking this action when in such a current state''. Since it uses a NN to generate actions, it is capable of dealing effectively RL tasks whose action space is continuous. The main difficulty is that a naive application of this actor-critic method with neural function approximators is unstable for challenging problems\cite{lillicrap2015continuous}. Hence, slowly-updating target NNs are introduced to ensure stability of training. The original DDPG algorithm is well presented in Algorithm 1 of Ref. \cite{lillicrap2015continuous}. In our work, we show that with the help of the continual-learning capability of a Bundle, we can effective stabilize the training, and can even get rid of the target NNs. In order to study the effect of being time-aware, we also introduce a bioclock for our modified DDPG algorithm. These result in a time-aware DDPG algorithm without target NNs (see Algorithm \ref{algo2}). 

The structures of the NNs are introduced in detail in the main text. For all models with a bioclock, we concatenate the output of the bioclock (i.e.,  $Y_{t}$ in Eq. \ref{eqn:bioclock}) with the conventional observation state $S_{t}$ to form a full input state vector. The settings of all the RL experiments are summarized in the Table \ref{tb1}. When making the figures, the curves for the testing loss are smoothened by a simple moving average whose window size is 100 for Fig 4a and 50 for other figures, and those for the training loss are smoothened by a simple moving average whose window size is 1000.
\begin{algorithm}
Randomly initialize a critic network $Q(s,t,a; W^{Q})$ with weights $W^{Q}$, an actor $\mu(s,t;W^{\mu})$ with weights $W^{\mu}$ (A biolock is included in each network)\\
Initialize a memory buffer $R$ for replay\\
\For{$episode=1, M$}{
Initialize a random process $\mathcal{N}$ for action exploration\\
Receive an initial observation state $s_{1}$\\
\For{$t=1, t_{m}$}{
Select an action $a_{t} = \mu(s_{t}, t; W^{\mu}) + \mathcal{N}_{t}$ according to the current policy and exploration noise\\
Execute action $a_{t}$ and obseve reward $r_{t}$ and the next state $s_{t+1}$\\
Store the transition $(s_t, a_t, r_t, s_{t+1}, t)$ in $R$\\
Sample a random minibatch of $N_{b}$ transitions $\{(s_{i}, a_{i}, r_{i}, s_{i+1}, t_{i})\}$ from $R$\\
Set $y_i = r_i + \gamma Q[s_{i+1}, t_{i}+1,\mu(s_{i+1}, t_{i}+1; W^{\mu}); W^{Q}]$\\
Update the critic by performing a gradient descent step to minimize the loss: $$L = \frac{1}{N_{b}} \sum_{i}[y_i - Q(s_i, t_i, a_i; W^{Q})]^2$$\\
Update the actor policy using the sampled policy gradient:
$$\nabla_{\theta^{\mu}}J \approx \frac{1}{N_{b}}\sum_{i}^{N_{b}}\nabla_{a} Q(s,t,a;W^{Q})|_{s=s_i, t=t_i, a=\mu(s_i, t_i)}\nabla_{W^{\mu}}\mu(s,t;W^{\mu})|_{s_i, t_i}$$

}
}
\caption{Time-aware DDPG without target NNs. To disable the time-awareness, we just need to omit the bioclocks and ignore the time labels. }\label{algo2}
\end{algorithm}

\begin{table}[]
\centering
\begin{tabular}{r|p{0.35\linewidth}|p{0.35\linewidth}}
& LunarLander-v2&Pendulum-v0\\
\hline
For all cases below&\makecell[l]{$t_m$=600\\fw-steps=50\\batch-size=512\\lr=5e-4\\$\gamma=0.99$\\n-evals=20\\eps-start=1.0\\ eps-end=0.01\\eps-decay=0.996\\ eval-interval=20\\n-runs=10}&\makecell[l]{memory-size=20000 (small)\\ 
or memory-size=200000 (large)\\$t_m$=600\\fw-steps=1\\batch-size=64\\ lr=5e-4\\$\gamma=0.99$\\n-evals=20\\ eval-interval=5\\n-runs=10}\\
\hline
MLP & hidden-size=573 (\# of trainable parameters is 8599)&hidden-size=860 (\# of trainable parameters is 16344)\\
\hline
MLP+Bioclock &  hidden-size=492 (\# of trainable parameters is 8594)&hidden-size=710 (\# of trainable parameters is 16416)\\
\hline
Bundle &  hidden-size=97 (\# of trainable parameters is 8587)&hidden-size=222 (\# of trainable parameters is 16318)\\
\hline
Bundle + Bioclock &  hidden-size=64 (\# of trainable parameters is 8578)&hidden-size=128 (\# of trainable parameters is 16290)\\
\hline
\end{tabular}
\caption{Settings of the RL experiments. $t_{m}$ is the maximum time step in an episode, fw-steps is the number of forward and backpropagation steps when training a batch, batch-size is the size of a batch, lr is the learning rate, $\gamma$ is the discount rate, n-evals is the number of repeated evaluations in one evaluation step, eval-interval is the number of episodes between two evaluation steps, n-runs is the number of repeated runs, the exploration probability $\epsilon$ (see Algorithm \ref{algo1}) for episode $i$ is given by $\epsilon(i)$=$\max($eps-end, eps-start * eps-decay$^{i})$, and memory-size is the size of the memory buffer for replay. In addition, we used an Ornstein-Uhlenbeck process with $\theta=0.15$ and $\sigma=0.2$\cite{lillicrap2015continuous} to generate the noise $\mathcal{N}_{t}$ in Algorithm \ref{algo2}; and $T_{min}$ and $T_{max}$ in Eq. \ref{eqn:bioclock} are 1 and $10^4$, respectively.}\label{tb1}
\end{table}
\end{document}